%% file: main.tex
\documentclass[smallabstract,smallcaptions]{dccpaper}
\usepackage{epsfig}
\usepackage{algorithm}
\usepackage{algorithmic}
\usepackage{citesort}
\usepackage{amsmath}
\usepackage{amssymb}
\usepackage{color}
\usepackage{url}
\usepackage{comment}
\usepackage{subcaption}
\usepackage{multirow}
\usepackage{tikz}

\newlength{\figurewidth}
\newlength{\smallfigurewidth}

\begin{document}

\title
{\large
\textbf{A Dual-Critic Reinforcement Learning Framework for Frame-level Bit Allocation in HEVC/H.265}
}

\author{%
Yung-Han Ho$^{\ast}$, Guo-Lun Jin$^{\ast}$, Yun Liang$^{\ast}$, Wen-Hsiao Peng$^{\ast}$,\\ and Xiaobo Li$^{\dag}$ \\[0.5em]
{\small\begin{minipage}{\linewidth}\begin{center}
\begin{tabular}{ccc}
$^{\ast}$National Chiao Tung University, Taiwan & \hspace*{0.5in} & $^{\dag}$Alibaba Group \\
\url{wpeng@cs.nctu.edu.tw} && \url{xiaobo.lixb@alibaba-inc.com}
\end{tabular}
\end{center}\end{minipage}}
}

\maketitle
\thispagestyle{empty}

\begin{abstract}
\input{Abstract}
\end{abstract}

\input{Intro}

\input{Method}

\input{Experiments}

\input{Conclusion}

\Section{References}
\bibliographystyle{IEEEbib}
\bibliography{refs}

\end{document}

%% file: Abstract.tex
This paper introduces a dual-critic reinforcement learning (RL) framework to address the problem of frame-level bit allocation in HEVC/H.265. The objective is to minimize the distortion of a group of pictures (GOP) under a rate constraint. Previous RL-based methods tackle such a constrained optimization problem by maximizing a single reward function that often combines a distortion and a rate reward. However, the way how these rewards are combined is usually ad hoc and may not generalize well to various coding conditions and video sequences. To overcome this issue, we adapt the deep deterministic policy gradient (DDPG) reinforcement learning algorithm for use with two critics, with one learning to predict the distortion reward and the other the rate reward. In particular, the distortion critic works to update the agent when the rate constraint is satisfied. By contrast, the rate critic makes the rate constraint a priority when the agent goes over the bit budget. Experimental results on commonly used datasets show that our method outperforms the bit allocation scheme in x265 and the single-critic baseline by a significant margin in terms of rate-distortion performance while offering fairly precise rate control.

%% file: Intro.tex
\section{Introduction}
\vspace{-0.5em}
Frame-level bit allocation is one key issue in video rate control \cite{chen2018reinforcement,hu2018reinforcement,kwong2020rate}. The task is to minimize the distortion of a group of pictures (GOP) under a rate constraint. In allocating a proper number of bits to every video frame in a GOP, it is crucial to consider the inter-frame dependencies because the quality of a coding frame depends highly on that of its reference frame. Thus, frame-level bit allocation within a GOP can be viewed as a dependent decision-making process with a long-term goal to achieve. 

Recently, deep reinforcement learning (RL) emerged as a promising technique for addressing dependent decision-making. Some early attempts apply RL to tackle constrained optimization problems in the video coding area~\cite{chen2018reinforcement,chung2017hevc,shi2020,hu2018reinforcement,kwong2020rate}. Chung et al.~\cite{chung2017hevc} utilized RL to determine the partition of coding tree units in HEVC/H.265. Hu et al.~\cite{hu2018reinforcement} cast the choice of quantization parameters for intra-frame coding as an RL problem. The idea was extended by Chen et al.~\cite{chen2018reinforcement} to frame-level bit allocation for inter-frame coding with hierarchical bi-prediction. Lately, Zhou et al.~\cite{kwong2020rate} took this line of research one step further for combined intra- and inter-frame bit allocation, focusing particularly on low-delay coding scenarios. Instead of optimizing video quality for human perception, Shi et al.~\cite{shi2020} proposed RL-based intra-frame bit allocation for image detection, classification, and segmentation.

In the context of frame-level bit allocation in a GOP, the reward is often formulated as a single function blending the resulting distortion of the GOP and the bias between the actual and target bit rates. Meeting the distortion and rate requirements simultaneously calls for a hyper-parameter, with the aim of seeking a balance between them.

To address this problem, we propose a dual-critic RL learning framework. We adapt the deep deterministic policy gradient (DDPG)~\cite{lillicrap2015continuous} algorithm for use with two critics: the rate and distortion critics. The rate critic learns to predict the rate bias upon the completion of encoding a GOP, while the distortion critic gives an estimate of the distortion-to-go along the coding process. Instead of using one signal reward function to guide the learning of an agent, which performs bit allocation by choosing a quantization parameter (QP) for every video frame, we apply both critics in an alternate way to train the agent. Extensive experimental results with x265 show that our method outperforms the bit allocation scheme in x265 and the single-critic method~\cite{chen2018reinforcement} by a significant margin in terms of rate-distortion (R-D) performance, when evaluated on Class B and Class C test sequences in JCT-VC dataset, which are not seen during training. It also offers fairly precise rate control accuracy. 


Our contributions include: (1) to the best of our knowledge, this is the first work that introduces a dual-critic-based RL framework to address constrained optimization problems; (2) we apply it to frame-level bit allocation for GOP coding with hierarchical bi-prediction; (3) our scheme outperforms the bit allocation scheme in x265 and the single-critic method \cite{chen2018reinforcement} in terms of R-D performance and rate control accuracy.

The remainder of this paper is organized as follows: Section~\ref{sec:PMethod} details our proposed method. Section~\ref{sec:Exp_results} presents the experimental results. Finally, Section~\ref{sec:Conclusion} concludes this work.

%% file: Method.tex
\vspace{-1em}
\section{Proposed Method}
\label{sec:PMethod}
\vspace{-0.5em}
\begin{figure}[t] 
\centering
\includegraphics[width=0.6\linewidth]{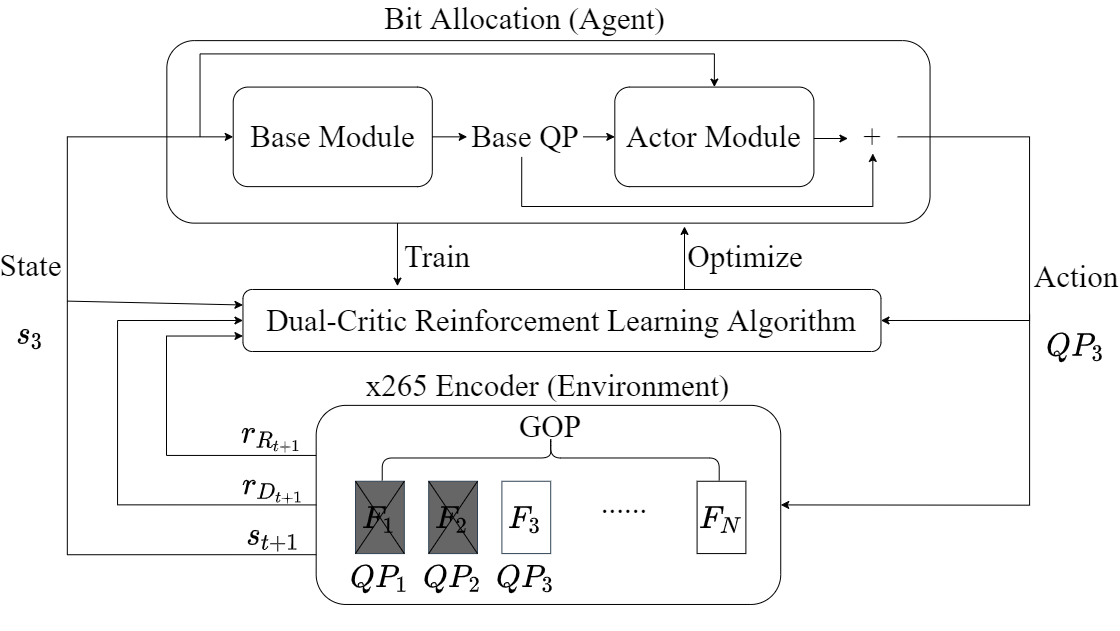}
\caption{The proposed RL framework for frame-level bit allocation.}
\label{fig:RL}
\vspace{-0.8em}
\end{figure}

\subsection{System Overview}
\label{subsec:Soverivew}

The frame-level bit allocation is to allocate a proper number of bits to every video frame by choosing a QP for its encoding, in order to minimize the distortion of a GOP subject to a rate constraint. In symbols, we have
\begin{equation}
\label{eq:goal}
\arg \min_{\{QP_i\}} \sum_{i=1}^{N}\ D_i(QP_i)~\text{s.t.} \sum_{i=1}^N R_i(QP_i)=R_{\text{GOP}},
\end{equation}
where $QP_i$ denotes the QP chosen for the $i$-th frame in a GOP, $D_i(QP_i)$ is the distortion incurred by encoding frame $i$ with $QP_i$, $R_i(QP_i)$ is the actual number of bits produced by the encoder, and $R_{\text{GOP}}$ is the GOP-level bit budget.

In this paper, we tackle the learning of an agent that can perform bit allocation among video frames in a GOP from an RL perspective. The frame-level bit allocation for a GOP is regarded as an episodic task, with the learning process given as follows and illustrated in Fig.~\ref{fig:RL}. (1) First, we evaluate a state signal (Section~\ref{subsec:Ssignals}) for an input frame. (2) Second, the state signal is fed to the agent implemented by a neural network (Section~\ref{subsec:Narchitecture}) to determine a QP for encoding the current frame. (3) Third, based on the chosen QP, the current frame is encoded and decoded by the specified codec (e.g.~x265) to update the state signal for the agent to decide the QP for the next video frame. Moreover, two different types of rewards (i.e.~the distortion and the rate rewards) are computed to indicate how good the choice of QP is (Section~\ref{subsec:DRrewards}). These steps are repeated iteratively until a terminal state (i.e.~the end of an episode) is reached. The training has the agent interact with the codec many times so that it can learn to maximize the rewards (Section~\ref{subsec:Dcritics}).

\vspace{-0.3em}
\subsection{State Signals}
\label{subsec:Ssignals}
\vspace{-0.3em}

The state signal serves as the input to the agent for its decision making. We follow mainly the state design in \cite{chen2018reinforcement}, which is summarized in Table~\ref{tab:state} for easy reference. It involves several hand-crafted intra- and inter-frame features that characterize the statistics of all the video frames in a GOP and their residual frames produced with zero motion vectors for simplicity. The fact that the state depends on the statistics of the remaining video frames in a GOP suggests a look-ahead strategy. As such, when a reference frame is not yet coded, we turn to its original and uncompressed version to compute the residual frame.

\begin{table*}[t]
\begin{center}
\centering
\caption{State Definition}
\vspace{-0.8em}
\label{tab:state}
\scalebox{0.75}{
\begin{tabular}{|c|c|}
\hline
\multicolumn{1}{c}{} & \multicolumn{1}{c}{state components}\\
\hline
\multicolumn{1}{l}{\textbf{1}} & \multicolumn{1}{l}{Intra-frame feature (the mean and variance of pixel values in a video frame)}  \\
\multicolumn{1}{l}{\textbf{2}} & \multicolumn{1}{l}{Inter-frame feature (the mean and variance of pixel values in a residual frame)}  
\\
\multicolumn{1}{l}{\textbf{3}} & \multicolumn{1}{l}{Average of intra-frame features over the remaining frames}  \\
\multicolumn{1}{l}{\textbf{4}} & \multicolumn{1}{l}{Average of inter-frame features over the remaining frames}  \\
\multicolumn{1}{l}{\textbf{5}} & \multicolumn{1}{l}{Percentage of the remaining bits} \\
\multicolumn{1}{l}{\textbf{6}} & \multicolumn{1}{l}{Number of remaining frames in the GOP} \\
\multicolumn{1}{l}{\textbf{7}} & \multicolumn{1}{l}{Temporal identification of the current frame} \\
\multicolumn{1}{l}{\textbf{8}} & \multicolumn{1}{l}{Target bit of the GOP} \\
\hline
\end{tabular}%
}
\end{center}
\vspace{-2em}
\end{table*}%

\vspace{-0.3em}
\subsection{Distortion and Rate Rewards}
\label{subsec:DRrewards}
\vspace{-0.3em}

The reward signal plays a central role in shaping the agent's behavior. The agent is usually trained to maximize a single reward. Because our task is to minimize the distortion of a GOP subject to a rate constraint, we develop two rewards: the distortion $r_D$ and the rate $r_R$ rewards. The former $r_D$ is evaluated upon the completion of encoding a video frame as an immediate reward, reflecting how good the current choice of QP is. It is defined as the normalized negative distortion of a compressed video frame in mean square error (MSE):   
\begin{equation}
\label{eq:Dreward}
r_{D_i} = -1 \times \frac{MSE_{i} - MSE_{QP0}}{(MSE_{QP51} - MSE_{QP0})\times GOP_{size}},
\end{equation}
where $MSE_i$ is the MSE of the current frame $i$, and the $MSE_{QP0}$ and $MSE_{QP51}$ are the MSE's of the entire GOP by encoding every video frame with the two extreme QP values, i.e. QP0 and QP51. The two normalization factors cause the sum of $r_{D_i}$ over the video frames in a GOP to fall approximately in the interval from 0 to 1. 
Unlike the distortion reward, which is evaluated for every video frame, the rate reward $r_R$ is given at the end of an episode, capturing the absolute bias between the actual bit rate and the target bit rate. It is specified by  
\begin{equation}
\label{eq:Rreward}
r_R = -1 \times \frac{|R_{GOP} - \sum_{i=1}^N R_i(QP_i)|}{R_{GOP}}.
\end{equation}

\vspace{-0.5em}
\subsection{Dual Critics}
\label{subsec:Dcritics}
\vspace{-0.3em}

To train the proposed agent following the aforementioned state and reward design, we depart from the single critic approach to learn two critics, with one $Q_D$ used for predicting the distortion reward (more precisely, the distortion-to-go $\sum_i r_{D_i}$) and the other $Q_R$ for predicting the rate reward $r_R$. The agent is updated adaptively with one of the critics. Specifically, we constantly apply the agent under training to encode a GOP. If it produces a bit rate exceeding the target $R_{GOP}$, we use the rate critic $Q_R$ to update its network parameters, with the aim of optimizing the agent towards precise rate control. In the other case, where the actual bit rate is lower than the target, it is updated with the distortion critic $Q_D$ to minimize the distortion of the reconstructed GOP.

Algorithm~\ref{alg:D_DDPG} details the training process. It can be divided into two major parts: Part I (lines 5 to 18) and Part II (lines 19 to 30). Part I corresponds to a rollout of the policy together with a noise process to collect transitions of states, actions, and rewards in a replay buffer $R_C$, the data of which are utilized to update the two critics $Q_D$ and $Q_R$ based on the ordinary DDPG algorithm. Note that a zero immediate rate reward is recorded at every step until  the very last step (the terminal step/state), at which the bit rate bias $r_R$ of the entire GOP is given as the immediate rate reward. Part II is another rollout of the policy but without the noise process. That is, the agent is put to use as for inference. The state transitions experienced is stored in a separate replay buffer $R_A$. At the end of the rollout, the actual bit rate used is compared against the target bit rate to decide which of the two critics should be used to update the agent.

\begin{algorithm}[t]
\begin{footnotesize}
\caption{The proposed dual-critic DDPG algorithm}
\label{alg:D_DDPG}
\begin{algorithmic}[1]
\STATE {Randomly initialize behavior critics $Q_D(s,a|w_D), Q_R(s,a|w_R)$, and behavior actor $\mu(s|\theta)$} with weights $w_D,w_R,$ and $\theta$
\STATE {Initialize target critics $Q_D'(s,a|w_D), Q_R'(s,a|w_R)$, and target actor $\mu'(s|\theta)$} with weights $w_D' \xleftarrow{} w_D,w_R' \xleftarrow{}w_R,$ and $\theta' \xleftarrow{}\theta$
\STATE {Initialize replay buffers $R_{C}$ and $R_{A}$}
\STATE {\textbf{for} episode = $1$ to $M$ \textbf{do}}
\STATE {\quad Initialize a random noise process $\mathcal{N}$ for action exploration}
\STATE {\quad Evaluate initial state $s_1$}
\STATE {\quad \textbf{for} frame $i = 1$ to $N$ in a GOP \textbf{do}}
\STATE {\quad\quad Set $a_i = \mu(s_i|\theta)+\mathcal{N}_i$ }
\STATE {\quad\quad Encode frame $i$ with $QP=a_i$}
\STATE {\quad\quad Evaluate immediate distortion reward $r_{D_i}$ and new state $s_{i+1}$}
\STATE {\quad\quad Evaluate immediate rate reward $r_{R_i} = (i == N)?r_R:0$}
\STATE {\quad\quad Store transition $(s_i, a_i, r_{D_i}, r_{R_i}, s_{i+1})$ in $R_{C}$}
\STATE {\quad \textbf{end for}}
\STATE {\quad Sample $N$ transitions $(s_n, a_n, r_{D_n}, r_{R_n}, s_{n+1})$ from $R_{C}$}
\STATE {\quad Set $y_{D_n} = r_{D_n}+\gamma Q_D'(s_{n+1},\mu'(s_{n+1}|\theta')|w_D')$ }
\STATE {\quad Update $Q_D$ by minimizing $L=\frac{1}{N}\Sigma_n{(y_{D_n}-Q_D(s_n,a_n|w_D))^2}$}
\STATE {\quad Set $y_{R_n} = r_{R_n}+\gamma Q_R'(s_{n+1},\mu'(s_{n+1}|\theta')|w_R')$ }
\STATE {\quad Update $Q_R$ by minimizing $L=\frac{1}{N}\Sigma_n{(y_{R_n}-Q_R(s_n,a_n|w_R))^2}$}
\STATE {\quad Evaluate initial state $s_1$ and set encoded bits $b=0$}
\STATE {\quad \textbf{for} frame $i = 1$ to $N$ in a GOP \textbf{do}}
\STATE {\quad\quad Set $a_i = \mu(s_i|\theta)$}
\STATE {\quad\quad Encode frame $i$ with $QP=a_i$}
\STATE {\quad\quad Evaluate new state $s_{i+1}$ and $b=b+R(QP)$}
\STATE {\quad\quad Store $s_i$ in $R_{A}$}
\STATE {\quad \textbf{end for}}
\STATE {\quad Evaluate terminal condition $B=(R_{GOP} < b)?1:0$}
\STATE {\quad Choose critic $Q=(B==1)? Q_{R}:Q_{D}$}
 \STATE {\quad Retrieve $s_n$ from $R_{A}$}
\STATE {\quad Update $\mu$ by maximizing $L=\frac{1}{N}\Sigma_n{Q(s_n,\mu(s_n|\theta))}$}
\STATE {\quad Clear replay buffer $R_{A}$}
\STATE {\quad Update $Q_D', Q_R',$ and $\mu'$ following DDPG}
\STATE {\textbf{end for}}
\end{algorithmic}
\end{footnotesize}
\end{algorithm}

\vspace{-0.3em}
\subsection{Network Architectures}
\label{subsec:Narchitecture}
\vspace{-0.3em}

As Fig.~\ref{fig:RL} illustrates, our agent consists of two modules: the base and the actor modules. The base module is pre-trained to mimic the QP control of the selected codec (e.g.~x265), while the actor module is learned with our dual-critic RL framework to update the QP of the base module by a delta QP in the interval $[-10,10]$. In training the actor module, the base module is freezed. The introduction of the base module restricts the exploration space for learning the actor module. This is found beneficial particularly because the convergence can be an issue for dual-critic training with a vast exploration space of QP options. The architectures of our actor and critic networks are detailed in Table~\ref{tab:architecture}. Note that the base module shares the same architecture as the actor module, except for the exclusion of the base QP as input (Fig.~\ref{fig:RL}).

\begin{table}[tb]
\centering

\caption{Architectures of the actor and critic networks.}
\vspace{-0.8em}
\label{tab:architecture}
\begin{footnotesize}
\scalebox{0.85}{
\begin{tabular}{|c|c||c|c|}
\hline

\multirow{1}{*}{\textbf{Layer}} &
\multicolumn{1}{c||}{\textbf{Actor}} & 
\multicolumn{2}{c|}{\textbf{Critics ($Q_D$ and $Q_R$)}} \\
\hline
\multirow{1}{*}{Input} & {State} & {State} & {Action} \\
\hline
\multirow{1}{*}{1} & {fc, 800, elu} & {fc, 500, leaky-relu} & {fc, 500, leaky-relu} \\
\hline
\multirow{1}{*}{2} & {fc, 500, elu} & {fc, 300, leaky-relu} & {fc, 300, none} \\
\hline
\multirow{1}{*}{3} & {fc, 1, sigmoid} & {fc, 300, none} & {-} \\
\hline
\multirow{1}{*}{4} & {-} & \multicolumn{2}{c|}{add, leaky-relu} \\
\hline
\multirow{1}{*}{5} & {-} & \multicolumn{2}{c|}{fc, 100, leaky-relu} \\
\hline
\multirow{1}{*}{6} & {-} & \multicolumn{2}{c|}{fc, 1, none} \\
\hline

\end{tabular}%
}
\end{footnotesize}
\end{table}

\vspace{-0.5em}
\subsection{Comparison with Single-Critic Design}
\label{subsec:SCritic}
\vspace{-0.5em}

Our scheme differs from the single-critic design \cite{chen2018reinforcement} in two aspects. First, it learns two separate critics, whereas Chen et al.~\cite{chen2018reinforcement} use one critic to learn a single reward function for a GOP as the weighted combination of the distortion and rate rewards, namely $-1 \times \sum_{i=1}^N r_{D_i} -  \lambda \times r_R$, where $\lambda$ is a hyper-parameter chosen empirically to trade the distortion of a GOP against the rate deviation from the target $R_{GOP}$. We argue that the use of a fixed $\lambda$, as adopted in \cite{chen2018reinforcement}, could hardly learn an agent that works well on different types of videos. Second, we require our agent to determine QP's for the remaining frames when the target $R_{GOP}$ is exceeded, whereas Chen et al.~\cite{chen2018reinforcement} propose encoding the remaining frames in terminal mode with $QP = QP_I + 10$, with $QP_I$ denoting the I-frame's QP.



%% file: Experiments.tex
\vspace{-0.5em}
\section{Experimental Results}
\label{sec:Exp_results}
\vspace{-1em}
\subsection{Settings}
We assess the objective and subjective compression performance of the proposed method, with the results compared against those produced by x265~\cite{x265}, the base module alone (referred hereafter to as base), and the single-critic method~\cite{chen2018reinforcement}. The baseline methods do not include \cite{kwong2020rate} because it is optimized specifically for low-delay p-frame coding, as compared to our GOP coding with hierarchical bi-prediction.  

\begin{figure}[t]
    \centering
    \includegraphics[width=0.75\linewidth]{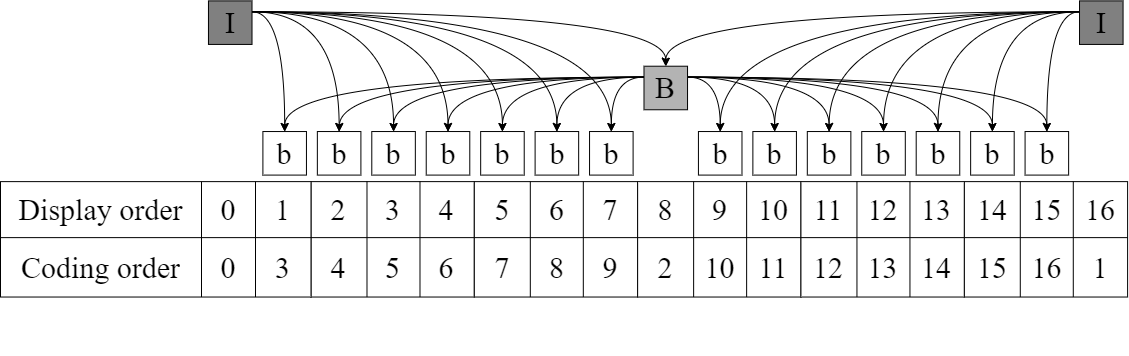}
    \vspace{-1.7em}
    \caption{The 3-level hierarchical bi-prediction structure.}
    \label{fig:coding_structure}
    \vspace{-0.3em}
\end{figure}

All the competing methods are evaluated on x265, with a hierarchical GOP coding structure as depicted in Fig.~\ref{fig:coding_structure}. Because our focus is on the bit allocation inside a GOP, all the tested methods follow the same GOP-level bit allocation as x265. This is achieved by first encoding every test sequence with fixed QP 22, 27, 32, and 37 to establish four sequence-level target bit rates $R_s$. They are then utilized as the sequence-level rate constraints to encode every test sequence again by turning on the ABR rate control of x265 (--bitrate $R_s$ --vbv-bufsize 2*$R_s$ --vbv-maxrate $R_s$), the results of which serve as our x265 baseline. For a fair comparison, the GOP-level bit rates produced by x265 in this ABR mode are taken as the GOP rate constraint $R_{GOP}$ for the other methods and the encoding is restricted to one pass only. 

We train separate models for the four target bit rates. The base module clones approximately the rate control behavior of x265 using its QP choices as the supervision ground truths. For a fair comparison, both our scheme and the single-critic method ~\cite{chen2018reinforcement} operate with the same base module. Our training datasets include UVG~\cite{10.1145/3339825.3394937}, MCL-JCV~\cite{7532610}, and Class A sequences in JCT-VC dataset. At test time, we use Class B and Class C sequences in JCT-VC dataset, which are not seen at training time. To speed up the RL training processes, the training and test videos are downscaled to $512 \times 320$. The run time of our feature extraction and actor network forwarding only occupies 4\% run time of ABR rate control of x265.

The objective compression performance is reported in terms of BD-PSNR gains and BD-rate savings, with x265 in ABR mode serving as anchor. The quality metrics are Y-PSNR and YUV-PSNR evaluated (and averaged) over individual frames, where YUV-PSNR is given by $(6 \times Y\_PSNR + U\_PSNR + V\_PSNR)/8$. The accuracy in rate control is quantified by averaging the absolute rate deviations from $R_{GOP}$ in percentage terms over all the GOPs in a test sequence.

\begin{figure}[t]
\begin{center}
\begin{subfigure}{0.3\textwidth}
    \centering
    \includegraphics[width=\linewidth]{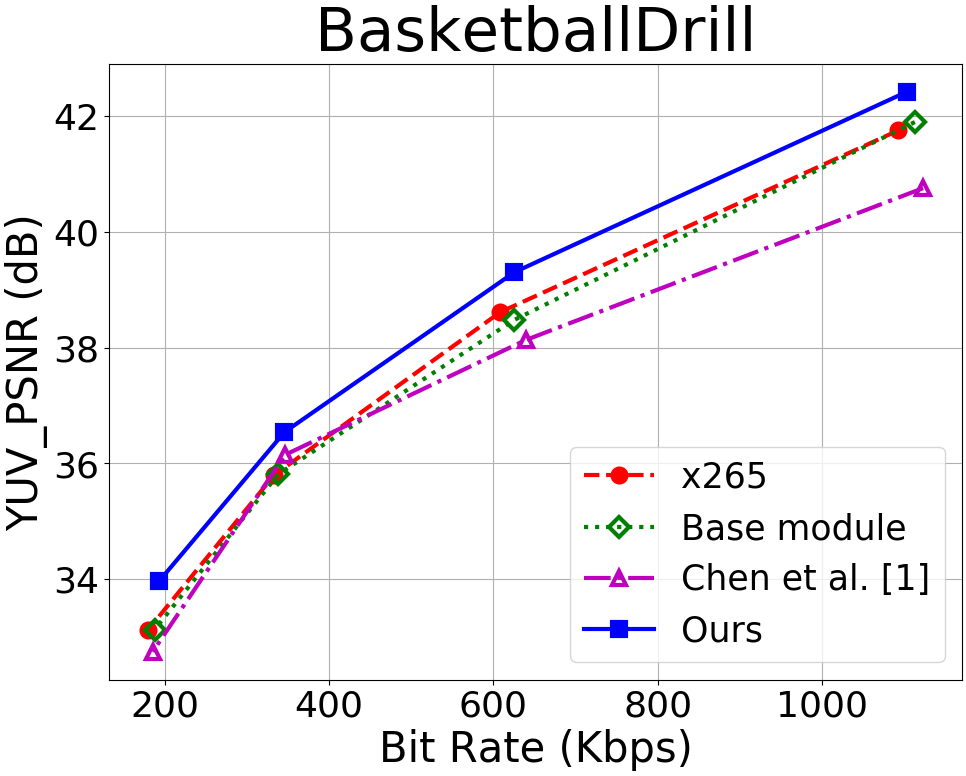} 
\end{subfigure}
\begin{subfigure}{0.3\textwidth}
    \centering
    \includegraphics[width=\linewidth]{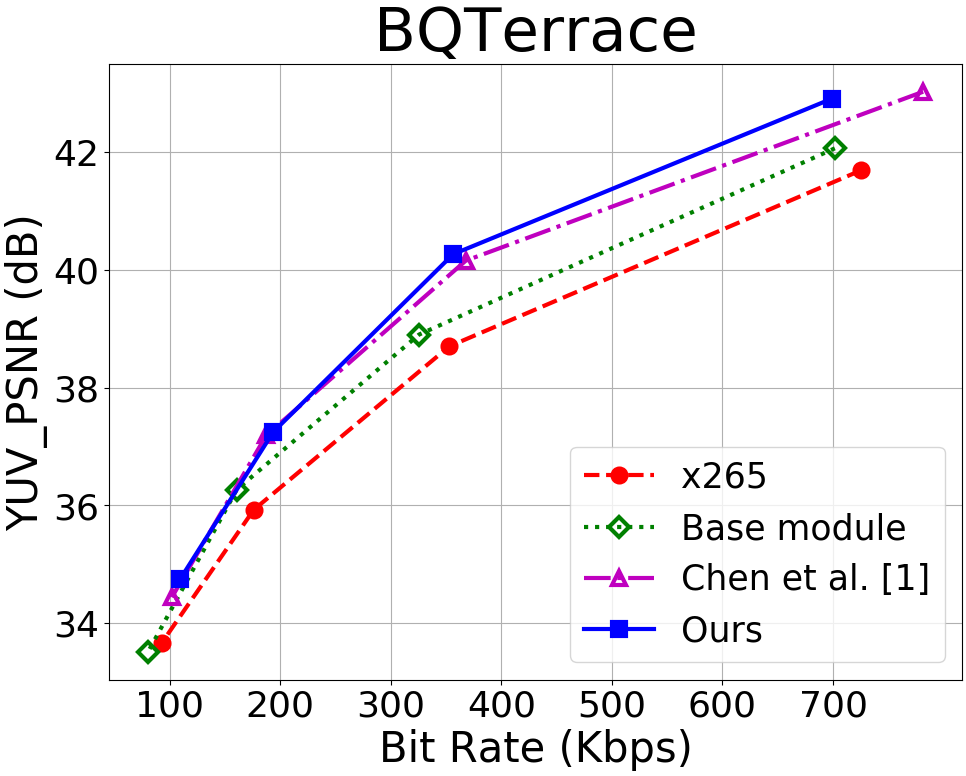}  
\end{subfigure}
\begin{subfigure}{0.3\textwidth}
    \centering
    \includegraphics[width=\linewidth]{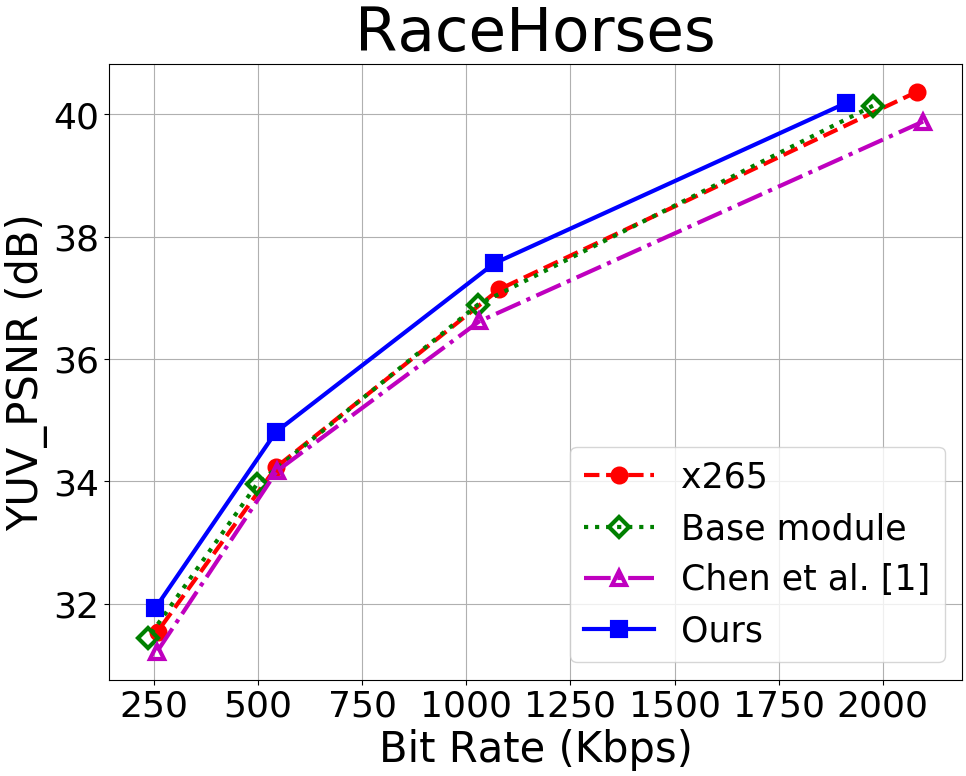}  
\end{subfigure}
\caption{Comparison of R-D curves for selected sequences.}
\label{fig:rd}
\end{center}
\vspace{-2.0em}
\end{figure}

\begin{table}[t]
\vspace{-0.8em}
\centering%
\caption{Comparison of BD-rates, BD-PSNRs, and rate deviations relative to x265}
\vspace{-0.8em}
\label{tab:exp_BD_PSNR}
\scalebox{0.6}{
    \begin{tabular}{|c|c|c|c|c|c|c|c|c|c|c|c|c|c|c|c|}
    \hline
    \multirow{3}{*}{\textbf{Sequences}} &
    \multicolumn{6}{c|}{\textbf{BD-rate (\%)}} &
    \multicolumn{6}{c|}{\textbf{BD-PSNR (dB)}} &
    \multicolumn{3}{c|}{\textbf{Rate deviation (\%)}}\\
    \cline{2-16}
    \multirow{1}{*}&
    \multicolumn{3}{c|}{Y}&
    \multicolumn{3}{c|}{YUV}&
    \multicolumn{3}{c|}{Y}&
    \multicolumn{3}{c|}{YUV}&
    \multicolumn{3}{c|}{-}\\
    \cline{2-16}
    &Base&[1]&Ours&Base&[1]&Ours&Base&[1]&Ours&Base&[1]&Ours&Base&[1]&Ours\\
    \hline
    BasketballDrill&3.2&6.6&\textbf{-9.9}&3.4&6.4&\textbf{-11.1}&-0.14&-0.39&\textbf{0.53}&-0.14&-0.37&\textbf{0.57}&7.7&5.6&\textbf{3.6}\\
    \hline
    BasketballDrive&-0.3&3.1&\textbf{-7.6}&-0.1&3.1&\textbf{-10.7}&0.02&-0.15&\textbf{0.40}&0.02&-0.15&\textbf{0.53}&11.3&\textbf{4.4}&5.0\\
    \hline
    BQMall&-7.9&-4.7&\textbf{-17.0}&-8.6&-5.5&\textbf{-22.7}&0.40&0.22&\textbf{0.92}&0.41&0.25&\textbf{1.18}&25.3&\textbf{9.6}&11.9\\
    \hline
    BQTerrace&-12.5&\textbf{-19.7}&-17.6&-13.3&-23.1&\textbf{-24.5}&0.59&\textbf{0.97}&0.88&0.57&1.07&\textbf{1.20}&16.8&\textbf{6.0}&7.7\\
    \hline
    Cactus&-4.7&-13.4&\textbf{-16.1}&-4.9&-15.5&\textbf{-19.8}&0.24&0.71&\textbf{0.85}&0.23&0.77&\textbf{0.99}&5.6&11.4&\textbf{4.4}\\
    \hline
    Kimono&-7.6&-16.3&\textbf{-17.4}&-7.2&-18.0&\textbf{-22.5}&	0.35&0.78&\textbf{0.84}&0.30&0.76&\textbf{0.98}&11.5&6.4&\textbf{4.9}\\
    \hline
    ParkScene&-2.1&-13.9&\textbf{-16.3}&-1.8&-15.8&\textbf{-20.9}&0.09&0.62&\textbf{0.73}&0.06&0.64&\textbf{0.87}&7.7&6.0&\textbf{4.9}\\
    \hline
    PartyScene&-10.1&-15.5&\textbf{-21.8}&-11.0&-17.0&\textbf{-25.9}&0.51&0.81&\textbf{1.19}&0.52&0.83&\textbf{1.33}&24.9&\textbf{5.8}&8.5\\
    \hline
    RaceHorses&-1.8&5.0&\textbf{-8.6}&-1.0&6.2&\textbf{-11.1}&0.08&-0.22&\textbf{0.41}&0.04&-0.25&\textbf{0.49}&18.6&6.1&\textbf{5.1}\\
    \hline
    \hline
    Average&-4.9&-7.6&\textbf{-14.7}&-5.0&-8.8&\textbf{-18.8}	&0.24&0.37&\textbf{0.75}&0.22&0.39&\textbf{0.91}&14.4&6.8&\textbf{6.2}\\
    \hline
    \end{tabular}
}
\end{table}

\begin{figure}[t]
\begin{subfigure}{0.30\textwidth}
    \centering
    \includegraphics[width=\linewidth]{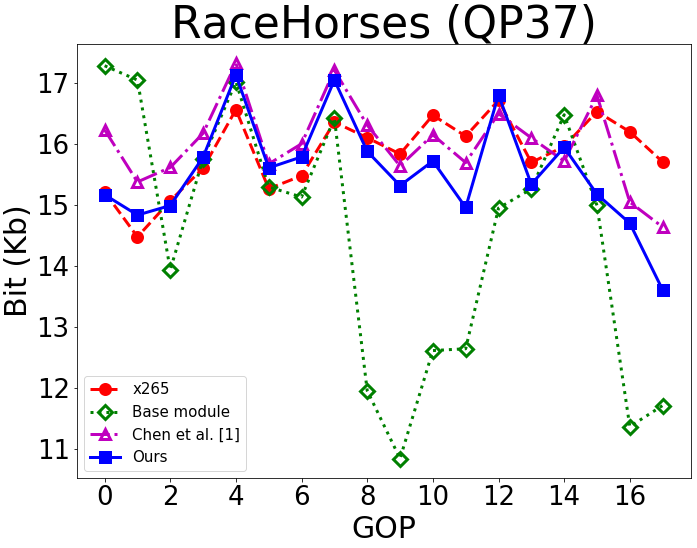} 
\end{subfigure}
\begin{subfigure}{0.70\textwidth}

    \centering
    \includegraphics[width=\linewidth]{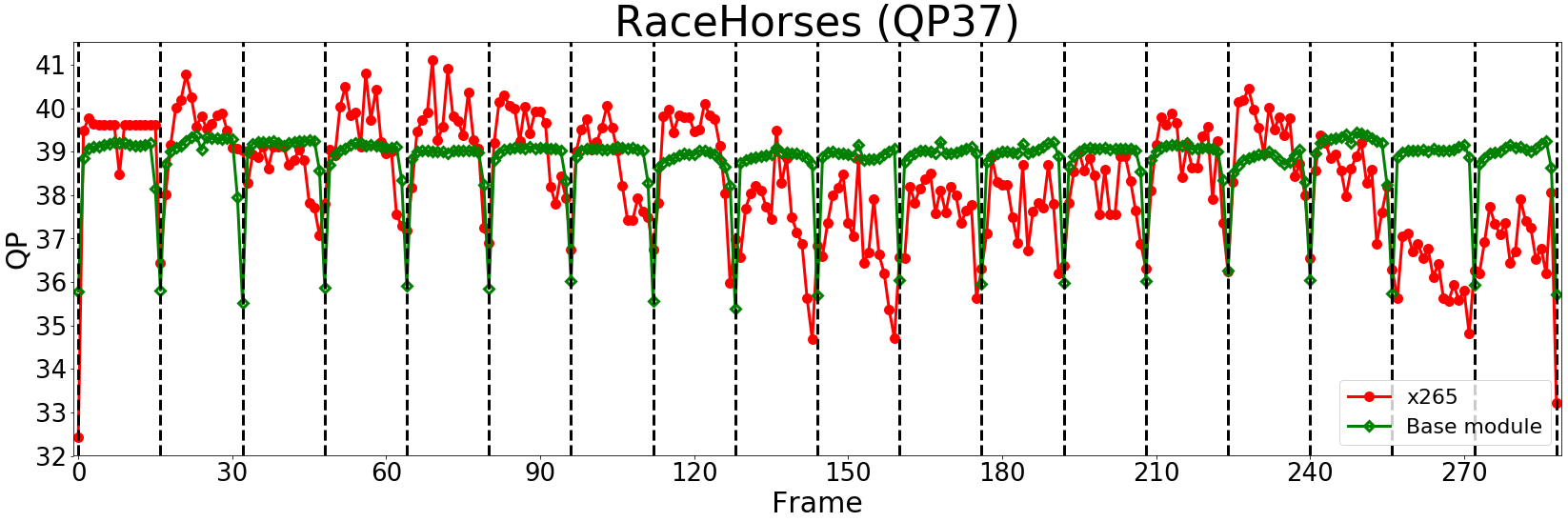} 
\end{subfigure}

\caption{Visualization of the coded GOP sizes for different methods (left) and the frame-level QP selection on RaceHorses for x265 and the base module (right).}
\label{fig:gop}
\vspace{-1em}
\end{figure}

\begin{figure}[t]
\begin{center}

\begin{subfigure}{0.3\textwidth}
    \centering
    \includegraphics[width=\linewidth]{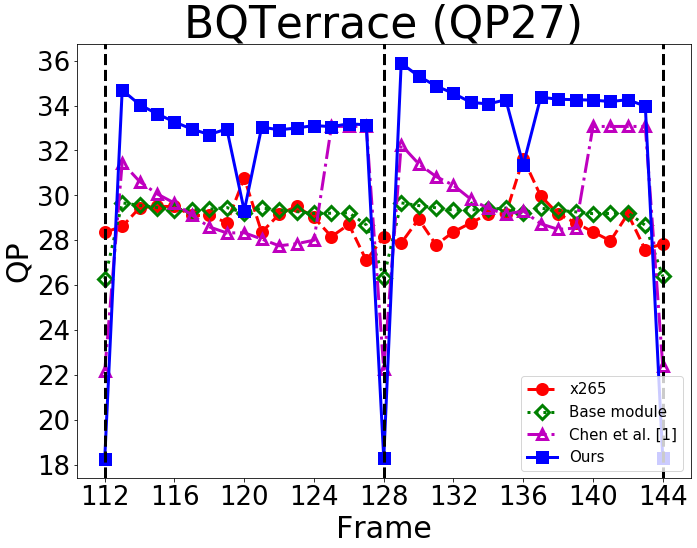}  
\end{subfigure}
\begin{subfigure}{0.3\textwidth}
    \centering
    \includegraphics[width=\linewidth]{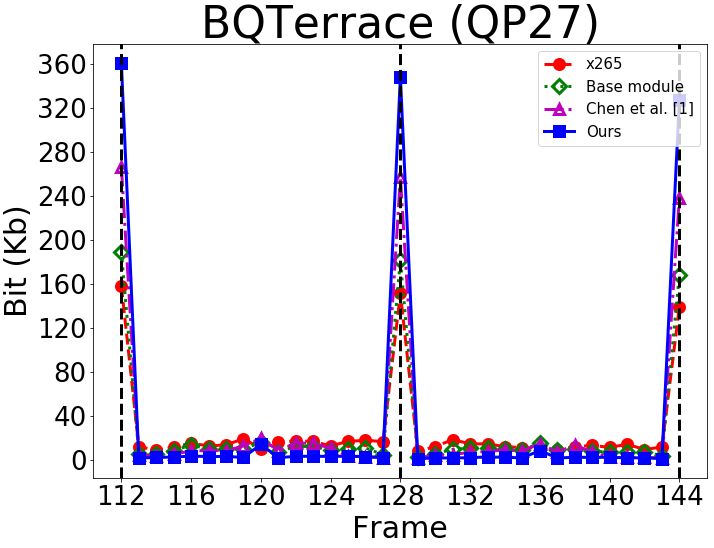}  
\end{subfigure}
\begin{subfigure}{0.3\textwidth}
    \centering
    \includegraphics[width=\linewidth]{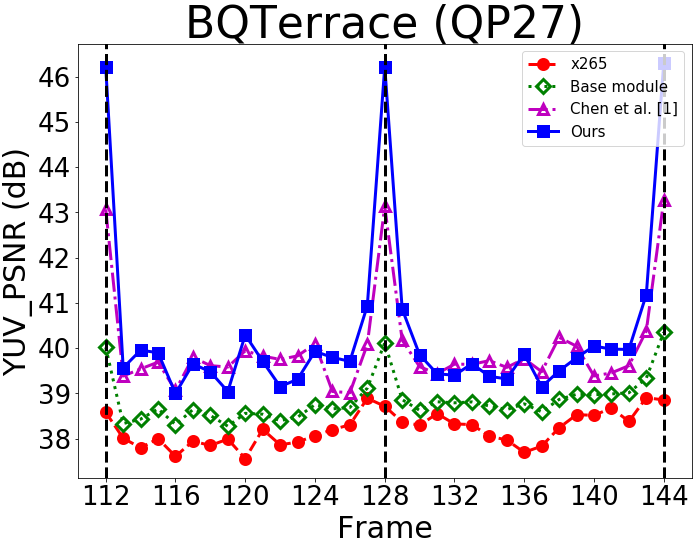}  
\end{subfigure}


\begin{subfigure}{0.3\textwidth}

    \centering
    \includegraphics[width=\linewidth]{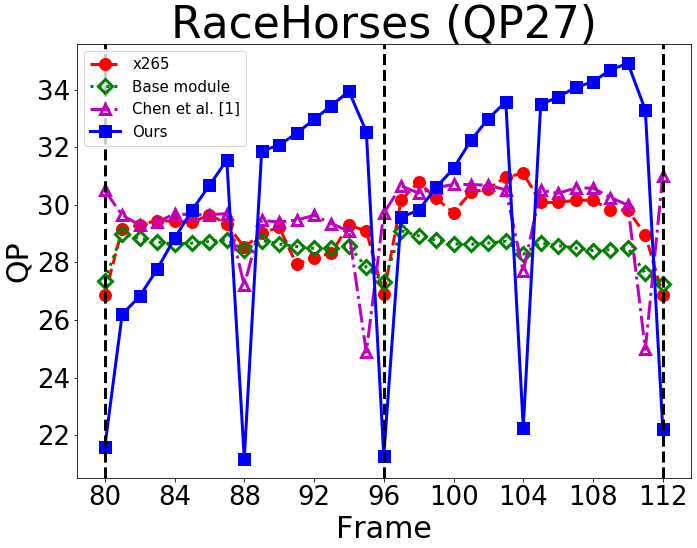}  
\end{subfigure}
\begin{subfigure}{0.3\textwidth}
    \centering
    \includegraphics[width=\linewidth]{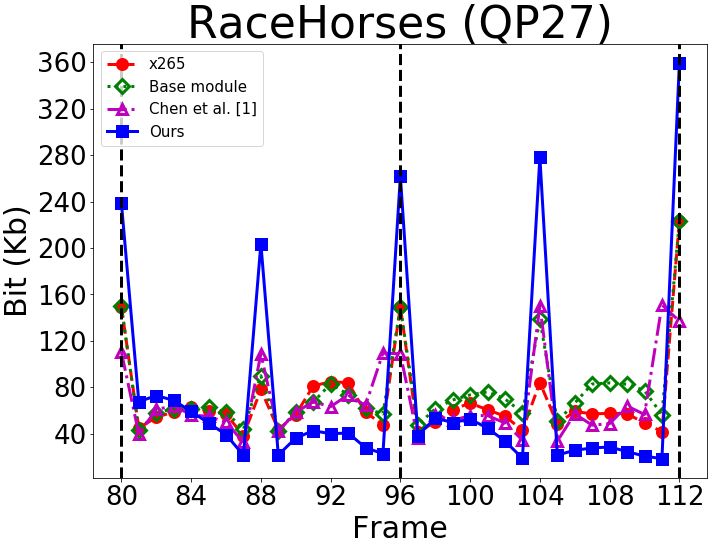}  
\end{subfigure}
\begin{subfigure}{0.3\textwidth}
    \centering
    \includegraphics[width=\linewidth]{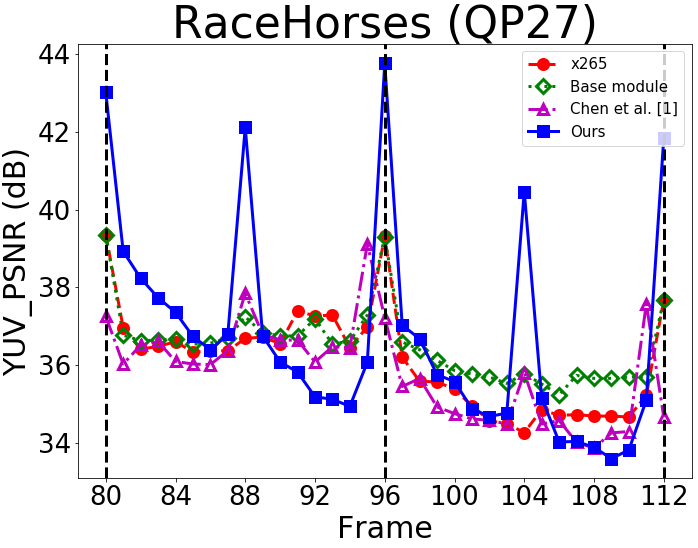}  
\end{subfigure}
\caption{Visualization of frame-level QP selection (left), coded frame size (middle), and YUV-PSNR (right) for two GOPs (frames 81-96 and frames 97-112).}
\label{fig:frame}
\end{center}
\vspace{-2em}
\end{figure}

\vspace{-0.5em}
\subsection{Compression Performance and Rate Control Accuracy}
\vspace{-0.5em}

Fig.~\ref{fig:rd} compares the rate-distortion (R-D) curves of different methods for three selected sequences, with the complete BD-rate and BD-PSNR results summarized in Table~\ref{tab:exp_BD_PSNR}. Also compared in the same table is the rate control accuracy averaged over all the rate points. From these results, the following observations are immediate: 




(1) \textit{Our method outperforms the competing methods in terms of R-D performance.} Fig.~\ref{fig:rd} shows that the proposed method achieves superior R-D performance to x265, as is confirmed by its BD-PSNR and BD-rate numbers in Table~\ref{tab:exp_BD_PSNR}. It improves Y-PSNR by 0.75dB and YUV-PSNR by 0.91dB, as compared to 0.37dB and 0.39dB improvements by Chen et al.~\cite{chen2018reinforcement}, respectively. The corresponding BD-rate savings range from 14.7\% to 18.8\%, which nearly double the rate savings by Chen et al.~\cite{chen2018reinforcement} (7.6\% to 8.8\%).




(2) \textit{The single-critic method has inconsistent R-D performance.} From Fig.~\ref{fig:rd}, it performs better than x265 in slow-motion sequences (e.g.~BQTerrace), but worse in the other fast-motion sequences (e.g.~BasketballDrill, RaceHorses). The performance degradation, if any, is more obvious at higher bit rates. Its inconsistent behavior in different types of videos and at different bit rates may be attributed to the use of a fixed hyper-parameter (Section~\ref{subsec:SCritic}) for trading off the video quality against the rate penalty, which is unable to generalize well to all the scenarios. 




(3) \textit{The base module is unable to meet the GOP-level rate constraints although showing comparable or better R-D performance than x265.} Table~\ref{tab:exp_BD_PSNR} reveals that its average deviation from the GOP-level bit rate constraint is around 14\% and can be as high as nearly 25\% in some sequences. However, with the RL training, both the single-critic method~\cite{chen2018reinforcement} and ours can match the GOP-level bit rates of x265 more closely (the left plot of Fig.~\ref{fig:gop}), showing much reduced rate deviations of 6.8\% and 6.2\% (Table~\ref{tab:exp_BD_PSNR}), respectively. From Table~\ref{tab:exp_BD_PSNR}, the base module achieves 4.9-5.0\% BD-rate savings and 0.22-0.24dB BD-PSNR gains over x265. This is because the ABR rate control may suffer from an unstable transient behavior in encoding the first few GOP's, as indicated by the developers of x265~\cite{x265}. By excluding these GOP's from behavior cloning, the base module learns a somewhat fixed and regular QP pattern over different GOP's, as shown in the right plot of Fig.~\ref{fig:gop}. Nevertheless, the higher rate control accuracy and much improved R-D performance of the proposed method than the base module stress the contributions of our RL framework (Table~\ref{tab:exp_BD_PSNR}).       

%


\vspace{-0.5em}
\subsection{Frame-level QP Assignment}
\vspace{-0.5em}
Fig.~\ref{fig:frame} visualizes the frame-level QP assignment within GOP's along with the corresponding coded frame size and YUV-PSNR. We make the following observations:



(4) \textit{Our RL agent tends to choose smaller QP values for I- and B-frames, allocating more bits to their coding.} In slow-motion sequences (e.g.~BQTerrace), our approach encodes I-frames at an even higher quality using an extremely small QP. Such a policy is intuitively agreeable since I- and B-frames serve as reference frames for the remaining non-reference b-frames. In slow-motion sequences, the quality of I-frames is even more critical to the reconstruction quality of a GOP.


(5) \textit{x265 and its clone version, i.e. the base module, focus more on I-frames without showing a drastic difference in QP selection for B- and b-frames.} A similar policy is observed in both fast- and slow-motion sequences. This causes B- and b-frames to have similar YUV-PSNR, and B-frames to use slightly more bits in fast-motion sequences (e.g.~RaceHorses) due to a longer-distance prediction. 

(6) \textit{The single-critic method \cite{chen2018reinforcement} learns a subtle policy on QP selection.} In slow-motion sequences (e.g.~BQTerrace), it behaves similarly to our scheme except that it does not particularly favor B-frames and enters terminal state early by encoding the remaining frames with $QP = QP_I + 10$ (Section~\ref{subsec:SCritic}). Interestingly, such a policy shows comparable R-D performance to ours (BQTerrace in Table~\ref{tab:exp_BD_PSNR}). However, in fast-motion sequences, I-frames are coded with relatively larger QP's than B- and b-frames, which explains its poorer R-D performance in these sequences.

(7) \textit{Our RL agent treats b-frames differently in fast-motion sequences.} It shows uneven QP assignment among b-frames, with their QP's increasing with the coding order (Fig.~\ref{fig:coding_structure}). Recall that our objective is to minimize the GOP-level distortion; there is no constraint on the frame-level quality distribution in a GOP. From Fig.~\ref{fig:coding_structure}, b-frames are coded independently of each other as non-reference frames. The QP assignment in Fig.~\ref{fig:gop} turns out to be a reasonably good solution since the GOP-level distortion is found to be smaller than those of the other competing methods.

\subsection{Subjective Quality Comparison}
\begin{figure}[t]
\centering
    
\includegraphics[width=\linewidth]{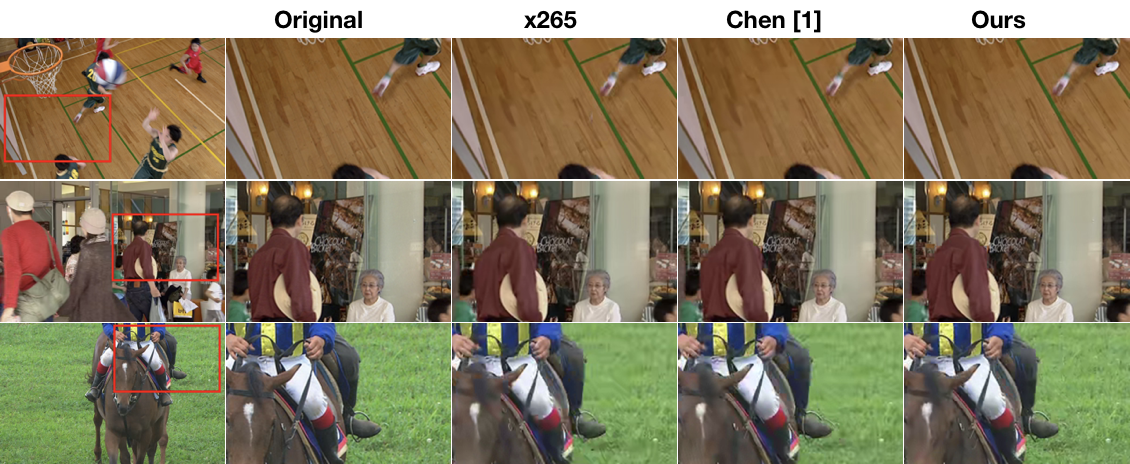}  
    
\caption{Subjective quality comparison of BasketballDrill (QP27, b-frame 407), BQMall (QP27, b-frame 125), and RaceHorses (QP37, b-frame 129).}
\label{fig:vision}
\vspace{-1em}
\end{figure}

Fig.~\ref{fig:vision} presents randomly chosen sample images produced by the competing methods. We observe that the proposed method retains more texture details, e.g. the wooden stripe in BasketballDrill, the crease of the shirt in BQMall, and the grass region in RaceHorses. Videos are provided at  {\url{http://mapl.nctu.edu.tw/RL_Rate_Control/}}.

%% file: Conclusion.tex
\vspace{-0.5em}
\section{Conclusion}
\label{sec:Conclusion}
\vspace{-0.5em}
This paper introduces a dual-critic-based RL framework for frame-level bit allocation in HEVC/H.265. It overcomes the need of combining the rate and distortion rewards in a heuristic manner with the single-critic design. The proposed method achieves promising R-D performance and fairly precise rate control accuracy. Although we are not currently aware of convergence guarantees for dual-critic training, it is able to arrive at a reasonably good solution in reality. This remains an open issue to be addressed in our future work.    


\vspace{-0.5em}
\section{Acknowledgements}
We are grateful to Yen-Kuang Chen and Minghai Qin for discussions and helpful feedback on the manuscript.
\vspace{-0.5em}